\documentclass{article} 
\usepackage[final]{nips_2016arxiv} 
\usepackage{hyperref}
\usepackage{url}

\usepackage{graphicx} 
\usepackage{subfigure} 

 \usepackage{array,multirow,graphicx}

\usepackage{natbib}
\usepackage{amssymb}
\usepackage{amsmath}
\usepackage{algorithm}
\usepackage{algorithmic}

\usepackage{framed}

\usepackage{hyperref}

\usepackage{xcolor}
\definecolor{shadecolor}{RGB}{150,150,150}

\definecolor{shadecolor}{RGB}{180,180,180}

\title{Dialog-based Language  Learning}

\author{Jason Weston \\
Facebook AI Research, \\
New York.\\
\texttt{jase@fb.com} \\ 
}

\newcommand{\PT}[1]{\textcolor{blue}{#1}}
\newcommand{\FT}[1]{\textcolor{black}{#1}}
\newcommand{\Z}{}
\newcommand{\PLUS}{{\textcolor{blue}{(+)}}}

\DeclareMathOperator*{\memn2n}{\text{MemN2N}}

\definecolor{dgreen}{rgb}{1.0,0,0.0}
\definecolor{dred}{rgb}{0.7,0.0,0.0}

\begin{document}

\maketitle

\begin{abstract}
A long-term goal of machine learning research is to build an intelligent dialog agent.
Most research in natural language understanding has focused on learning
from fixed training sets of labeled data, with supervision either at the word level
(tagging, parsing tasks) or sentence level (question answering, machine translation).
This kind of supervision is not realistic of how humans learn, where language is both
learned by, and used for, communication.
In this work, we study dialog-based language learning, where supervision 
 is given naturally and implicitly in the response of the dialog partner during 
the conversation.
%
We study this setup in two domains: 
the bAbI dataset of \cite{weston2015towards} and large-scale                              
question answering from \cite{dodge2015evaluating}. 
We evaluate a set of baseline learning strategies
 on these tasks, and show that a novel model
incorporating predictive lookahead is a promising 
approach  for learning from a teacher's response.
In particular, a surprising result is that it can learn to answer questions correctly without any reward-based supervision at all.
\end{abstract}

\section{Introduction}

Many of machine learning's successes have come from supervised learning, which
typically involves employing annotators to label large quantities of data per task.
However, humans can learn  by acting and learning from the consequences of (i.e, the feedback from) 
their actions.
When humans act in dialogs (i.e., make speech utterances)
 the feedback is from other human's responses, which hence contain very rich information.
This is perhaps most pronounced in a student/teacher scenario where the 
teacher provides positive feedback for
successful communication and corrections for
unsuccessful ones \cite{latham1997learning,werts1995instructive}.
However, in general any reply from a dialog partner, teacher or not,
is likely to contain an informative training signal for learning how to use language in subsequent conversations.

In this paper we explore whether we can train machine learning models 
to learn from dialogs.
The ultimate goal is to be able to develop an intelligent dialog agent 
that can learn {\em while conducting conversations}. To do that it needs to learn from feedback
that is supplied as natural language.
However, most machine learning tasks in the natural language processing literature are not of this form:
 they are either hand labeled at the word level (part of speech tagging, named entity recognition),
segment (chunking) or sentence level (question answering) by labelers.
Subsequently, learning algorithms have been developed to learn from that kind of supervision.
We therefore need to develop evaluation datasets for the dialog-based language learning setting,
as well as developing models and algorithms able to learn in such a regime.

The contribution of the present work is thus:
\begin{itemize}
\item We introduce a set of tasks that model natural feedback from a teacher and hence assess the feasibility of dialog-based language learning.
\item We evaluate some baseline models on this data, comparing to standard supervised learning.
\item We introduce a novel forward prediction model, whereby the learner tries to predict the teacher's replies to its actions,
yielding promising results, even with no reward signal at all.
\end{itemize}

\section{Related Work}

In human language learning
the usefulness of social interaction and
 natural infant directed conversations
is emphasized, see e.g. the review paper \cite{kuhl2004early},
although the usefulness of feedback for learning grammar is disputed
\cite{marcus1993negative}. Support for the usefulness of feedback is found however 
in second language learning  \cite{bassiri2011interactional} and
 learning by students \cite{higgins2002conscientious,latham1997learning,werts1995instructive}.

In machine learning, one line of research has focused on supervised learning from
dialogs using neural models \cite{sordoni2015neural,dodge2015evaluating}.
Question answering given either a database of knowledge \cite{bordes2015large} or
short stories \cite{weston2015towards} can be considered as a simple case of dialog
which is easy to evaluate. Those tasks typically do not consider feedback.
There is  work on the 
the use of feedback and dialog for
learning, notably for 
collecting knowledge  to answer questions \cite{hixon2015learning,pappu2013predicting},
the use of natural language instruction for learning symbolic rules
 \cite{kuhlmann2004guiding,goldwasser2014learning} and the use of binary feedback (rewards) for learning parsers \cite{clarke2010driving}.

Another setting which  uses feedback is the setting of reinforcement learning, 
see e.g. \cite{rieser2011reinforcement, schatzmann2006survey} for a summary of its use in dialog.
However, those approaches often
consider reward as the feedback model rather than exploiting the dialog feedback per se.
Nevertheless, reinforcement learning ideas have been used to good effect for other tasks as
well, such as understanding text adventure games \cite{narasimhan2015language}, 
image captioning \cite{xu2015show},  machine translation and summarization  \cite{ranzato2015sequence}. 
Recently, \cite{mikolov2015roadmap} also proposed a reward-based learning framework for
learning how to learn.

Finally, forward prediction models, which we make use of in this work, have been used for
learning eye tracking \cite{schmidhuber1991learning},
controlling robot arms \cite{lenz2015deepmpc} and vehicles \cite{wayne2014hierarchical},
and action-conditional video prediction in atari games \cite{oh2015action,stadie2015incentivizing}.
We are not aware of their use thus far for dialog.

\begin{figure*}[t]
\begin{small}
\caption{Sample dialogs with differing supervision signals (tasks 1 to 10).
In each case the same example is given for simplicity.
Black text is spoken by the teacher, red text denotes responses by the learner,
blue text is provided by an expert student (which the learner can imitate),
\PLUS~denotes positive reward external to the dialog (e.g. feedback provided by another medium, such as a nod of the head from the teacher). 
  \label{taskseta}}\vspace{2mm}
\begin{tabular}{|l|c|l|}
\cline{1-1}\cline{3-3}
&& \\[-2ex]
{\bf Task 1: Imitating an Expert Student}    &&   {\bf Task 2: Positive and Negative Feedback} \\
&& \\[-2ex]
~~Mary went to the hallway.                   &&~~Mary went to the hallway.  \\
~~John moved to the bathroom.                  &&~~John moved to the bathroom.\\
~~Mary travelled to the kitchen.               &&~~Mary travelled to the kitchen.\\
~~Where is Mary? \textcolor{blue}{A:kitchen} &&~~Where is Mary?  \textcolor{dgreen}{A:playground} \Z \\
~~Where is John?  \textcolor{blue}{A:bathroom}    &&~~No, that's incorrect. \Z \\
 &&~~Where is John?  \textcolor{dgreen}{A:bathroom} \Z \\
 &&~~Yes, that's right! \PLUS \\
\cline{1-1}\cline{3-3} 
\multicolumn{3}{c}{}
\vspace{-2mm}
\\
\cline{1-1}\cline{3-3}
&& \\[-2ex]
{\bf Task 3: Answers Supplied by Teacher}   &&   {\bf Task 4: Hints Supplied by Teacher} \\
&& \\[-2ex]
~~Mary went to the hallway.                   &&~~Mary went to the hallway.  \\
~~John moved to the bathroom.                   &&~~John moved to the bathroom.\\
~~Mary travelled to the kitchen.               &&~~Mary travelled to the kitchen.\\
~~Where is Mary?  \textcolor{dgreen}{A:bedroom} \Z  &&~~Where is Mary?  \textcolor{dgreen}{A:bathroom} \Z  \\
~~No, the answer is kitchen. \Z 
                              &&~~No, they are downstairs. \Z \\
 ~~Where is John?  \textcolor{dgreen}{A:bathroom} \Z &&  ~~Where is John?  \textcolor{dgreen}{A:kitchen} \Z   \\
 ~~Correct! \PLUS && ~~No, they are upstairs. \Z  \\
\cline{1-1}\cline{3-3} \multicolumn{3}{c}{} \vspace{-2mm} \\ 
\cline{1-1}\cline{3-3} && \\[-2ex]
{\bf Task 5: Supporting Facts Supplied by Teacher}   &&  {\bf Task 6: Partial Feedback}    \\
&& \\[-2ex]
~~Mary went to the hallway.                   &&~~Mary went to the hallway.  \\
~~John moved to the bathroom.                  &&~~John moved to the bathroom.\\
~~Mary travelled to the kitchen.               &&~~Mary travelled to the kitchen.\\
~~Where is Mary?  \textcolor{dgreen}{A:kitchen} \Z  &&~~Where is Mary?  \textcolor{dgreen}{A:kitchen} \Z  \\
~~Yes, that's right! \PLUS       && ~~Yes, that's right!  \\

~~Where is John?  \textcolor{dgreen}{A:hallway} \Z && ~~Where is John?  \textcolor{dgreen}{A:bathroom} \Z \\

~~No, because John moved to the bathroom.   \Z &&  ~~Yes, that's correct!  \PLUS \\ 
\cline{1-1}\cline{3-3} \multicolumn{3}{c}{} \vspace{-2mm} \\ 
\cline{1-1}\cline{3-3} && \\[-2ex]
{\bf Task 7: No Feedback}          &&   {\bf Task 8: Imitation and Feedback Mixture} \\
&& \\[-2ex]
~~Mary went to the hallway.                   &&~~Mary went to the hallway.  \\
~~John moved to the bathroom.                  &&~~John moved to the bathroom.\\
~~Mary travelled to the kitchen.               &&~~Mary travelled to the kitchen.\\
~~Where is Mary?  \textcolor{dgreen}{A:kitchen}   &&~~Where is Mary?  \textcolor{blue}{A:kitchen}  \\
~~Yes, that's right!  
                                              &&~~Where is John?  \textcolor{dgreen}{A:bathroom} \Z \\ 
~~Where is John?  \textcolor{dgreen}{A:bathroom}  &&  ~~That's right!  \PLUS \\ 
~~Yes, that's correct!  && \\ 
\cline{1-1}\cline{3-3} \multicolumn{3}{c}{} \vspace{-2mm} \\ 
\cline{1-1}\cline{3-3} && \\[-2ex]
{\bf Task 9:  Asking For Corrections}      &&   {\bf Task 10: Asking For Supporting Facts}  \\
&& \\[-2ex]
~~Mary went to the hallway.                   &&  ~~Mary went to the hallway.    \\
~~John moved to the bathroom.                  && ~~John moved to the bathroom.  \\
~~Mary travelled to the kitchen.               && ~~Mary travelled to the kitchen. \\
~~Where is Mary?  \textcolor{dgreen}{A:kitchen} \Z  && ~~Where is Mary?  \textcolor{dgreen}{A:kitchen} \\
~~Yes, that's right! \PLUS       && ~~Yes, that's right! \PLUS   \\
~~Where is John?  \textcolor{dgreen}{A:hallway} \Z && ~~Where is John?  \textcolor{dgreen}{A:hallway} \Z  \\
~~No, that's not right.   \textcolor{dgreen}{A:Can you help me?} \Z && ~~No, that's not right.   \textcolor{dgreen}{A:Can you help me?} \Z \\
~~Bathroom. \Z  &&  ~~A relevant fact is John moved to the bathroom.  \Z \\
\cline{1-1}\cline{3-3}
\end{tabular}
\end{small}
\vspace*{-3ex}
\end{figure*}

\begin{figure*}[t]
\begin{small}
\caption{Samples from the MovieQA dataset \cite{dodge2015evaluating}.
In our experiments we consider 10 different language learning setups as described in Figure \ref{taskseta} and Sec.  \ref{sec:tasks}. The examples given here are for tasks 2 and 3,
questions are in black and answers in red, and $\PLUS$                                       
indicates receiving positive reward.    
 \label{taskset2}}\vspace{2mm}
\begin{tabular}{|l|c|l|}
\cline{1-1}\cline{3-3}
&& \\[-2ex]
{\bf Task 2: Positive and Negative Feedback} &&
{\bf Task 3: Answers Supplied by Teacher}  \\
&& \\[-2ex]
What movies are about open source?   \textcolor{dred}{Revolution OS}       &&
What films are about Hawaii?  \textcolor{dred}{50 First Dates}\\
That's right! \PLUS &&
Correct! \PLUS \\
What movies did Darren McGavin star in?       \textcolor{dred}{Carmen} &&
Who acted in Licence to Kill?  \textcolor{dred}{Billy Madison}\\
Sorry, that's not it.                &&
No, the answer is Timothy Dalton.\\
Who directed the film White Elephant?  \textcolor{dred}{M. Curtiz} &&
What genre is Saratoga Trunk in?      \textcolor{dred}{Drama}\\  
No, that is incorrect. &&
Yes! \PLUS \\
\cline{1-1}\cline{3-3}
\end{tabular}
\end{small}
\end{figure*}

\section{Dialog-Based Supervision Tasks} \label{sec:tasks}

Dialog-based supervision comes in many forms. As far as we are aware it is a currently unsolved
problem 
which type of learning strategy will work in which setting. 
In this section we therefore identify different modes of dialog-based supervision, and build
a learning problem for each. The goal is to then evaluate learners
on each type of supervision.

We thus begin by selecting two existing  datasets:
(i) the single supporting fact problem from the bAbI datasets
\cite{weston2015towards} which consists of short stories from a simulated world
followed by questions; and (ii) the MovieQA dataset \cite{dodge2015evaluating}
which is a large-scale dataset ($\sim100k$ questions over $\sim75k$ entities)
based on questions with answers in the open movie database (OMDb). 
For each dataset
we then consider ten modes of dialog-based supervision.
The supervision modes are summarized in Fig. \ref{taskseta} using a snippet of
the bAbI dataset as an example.
The same setups are also used 
for MovieQA,  some examples of which are given in Fig \ref{taskset2}.

We now describe each supervision setup in turn.

\paragraph{Imitating an Expert Student}
In Task 1 the dialogs take place between a teacher and an expert student  who gives
semantically coherent answers. Hence, the task is for the learner to imitate that expert 
student, and become an expert themselves.
For example, imagine the real-world scenario where
a child observes their two parents talking to each other, it can learn but it
is not actually taking part in the conversation.
Note that our main goal in this paper
 is to examine how a non-expert can learn to improve its dialog skills while conversing.
The rest of our tasks will hence concentrate on that goal. This task can be seen as a natural
baseline for the rest of our tasks given the same input dialogs and questions.

\paragraph{Positive and Negative Feedback} 
In Task 2, when the learner answers a question the teacher then replies with either positive
or negative feedback. In our experiments the 
subsequent responses are variants of ``No, that's incorrect'' or ``Yes, that's right''.
In the datasets we build there are 6 templates for positive feedback and 6 templates for negative
feedback, e.g. "Sorry, that's not it.'', ''Wrong'', etc.
To separate the notion of positive from negative (otherwise the signal is just words with no 
notion that yes is better than no) we assume an additional {\em external reward} signal that is not part of the text.
As shown in Fig. \ref{taskseta} Task 2, $\PLUS$
denotes positive reward external to the dialog (e.g. feedback provided by another medium, 
such as a nod of the head from the teacher). This is provided with every positive response.
Note the difference in supervision compared to Task 1: there
every answer is right and provides positive supervision. Here, only the answers
the learner got correct have positive supervision. 
This could clearly be a problem when the learner is unskilled: it will supply incorrect answers and
never (or hardly ever) receive positive responses.

\paragraph{Answers Supplied by Teacher}
In Task 3 the teacher gives positive and negative feedback as in Task 2, however
 when the learner's answer is incorrect, the teacher also responds with the correction.
For example  if ``where is Mary?'' is answered with the incorrect answer
``bedroom'' the teacher
responds ``No, the answer is kitchen''', see Fig. \ref{taskseta} Task 3.
If the learner knows how to use this extra information, it effectively has as much supervision signal as
with Task 1, and much more than for Task 2.

\paragraph{Hints Supplied by Teacher}
In Task 4, the corrections provided by the teacher do not provide the exact answer
as in Task 3, but only a useful hint.
This setting is meant to mimic the real life occurrence of being provided only partial
 information about what you did wrong. In our datasets we do this by providing the {\em class}
of the correct answer, e.g. ``No, they are downstairs'' if the answer should be kitchen,
or ``No, it is a director'' for the question ``Who directed Monsters, Inc.?'' (using OMDB metadata).
The supervision signal here is hence somewhere in between Task 2 and 3.

\paragraph{Supporting Facts Supplied by Teacher}
In Task 5, another way of providing partial supervision for an incorrect answer is explored.
Here, the teacher gives a reason (explanation) 
why the answer is wrong by referring to a known fact that supports the true answer that the
incorrect answer may contradict. For example ``No, because John moved to the bathroom'' for
an incorrect answer to ``Where is John?'', see Fig. \ref{taskseta} Task 5.
This is related to 
 what is termed {\em strong supervision}  in \cite{weston2015towards} where supporting facts
and answers are given for question answering tasks.

\paragraph{Partial Feedback} Task 6 considers the case where external rewards are only given some
of (50\% of) the time for correct answers, the setting is otherwise identical to Task 3.
This attempts to mimic the realistic situation of some
learning being more closely supervised (a teacher rewarding you for getting some answers right) whereas
other dialogs have less supervision (no external rewards). 
The task attempts to assess the impact of such partial supervision.

\paragraph{No Feedback} In Task 7 external rewards are not given at all,
only text, but is otherwise identical to Tasks 3 and 6.
This task explores whether it is actually possible to learn how to answer at all in such 
a setting. We find in our experiments the answer is surprisingly yes, at least in some conditions. 

\paragraph{Imitation and Feedback Mixture} 
Task 8 combines Tasks 1 and 2. The goal is to see if a learner can learn successfully from both
forms of supervision at once. This mimics a child both observing pairs of experts talking (Task 1) 
while also trying to talk (Task 2).

\paragraph{Asking For Corrections} Another natural way of collecting supervision is for the learner
to ask questions of the teacher about what it has done wrong. Task 9 tests one of the most simple 
instances, where asking ``Can you help me?'' when wrong obtains from the teacher the correct
answer. This is thus
related to the supervision in Task 3 except the learner must first ask for help in the dialog.
This is potentially harder for a model as the relevant information is spread over a larger context.

\paragraph{Asking for Supporting Facts} Finally, in Task 10,
 a second less direct form of supervision for
the learner 
after asking for help is to receive a hint rather than the correct answer, such
as ``A relevant fact is John moved to the bathroom'' when asking ``Can you help me?'', see Fig. \ref{taskseta} Task 10. This is thus
related to the supervision in Task 5 except the learner must request help.

\vspace{3mm}

In our experiments  we constructed the ten supervision tasks for the two
datasets which are all available for download at \url{http://fb.ai/babi}.
They were built in the following way: for each task we consider a fixed policy\footnote{
Since the policy is fixed and actually does not depend on the model being learnt, one could also think of it as coming from another agent (or the same agent in the past) which in either case is an imperfect expert.}
%
for performing actions
(answering questions) which gets questions correct with probability $\pi_{acc}$
(i.e. the chance of getting the red text correct in Figs. \ref{taskseta} and \ref{taskset2}).
We thus can compare different learning algorithms
for each task over different values of $\pi_{acc}$ (0.5, 0.1 and 0.01).
In all cases  a training, validation and test set is provided.
For the bAbI dataset this consists of 1000, 100 and 1000 questions respectively per task,
and for movieQA there are  $\sim96k$, $\sim10k$ and $\sim10k$ respectively.
MovieQA also includes a knowledge base (KB) of $\sim85k$ facts from OMDB,
the memory network model we employ uses inverted index retrieval based on the question
to form relevant memories from this set,  see \cite{dodge2015evaluating} for more details.
Note that because the policies are fixed the experiments in this paper
are {\em not} in a reinforcement learning setting. 

\begin{figure*}[h!]
\caption{Architectures for (reward-based) imitation and forward prediction.
\label{fig:archs}
}
\begin{center}
\vspace{-7mm}
\includegraphics[width=7.3cm]{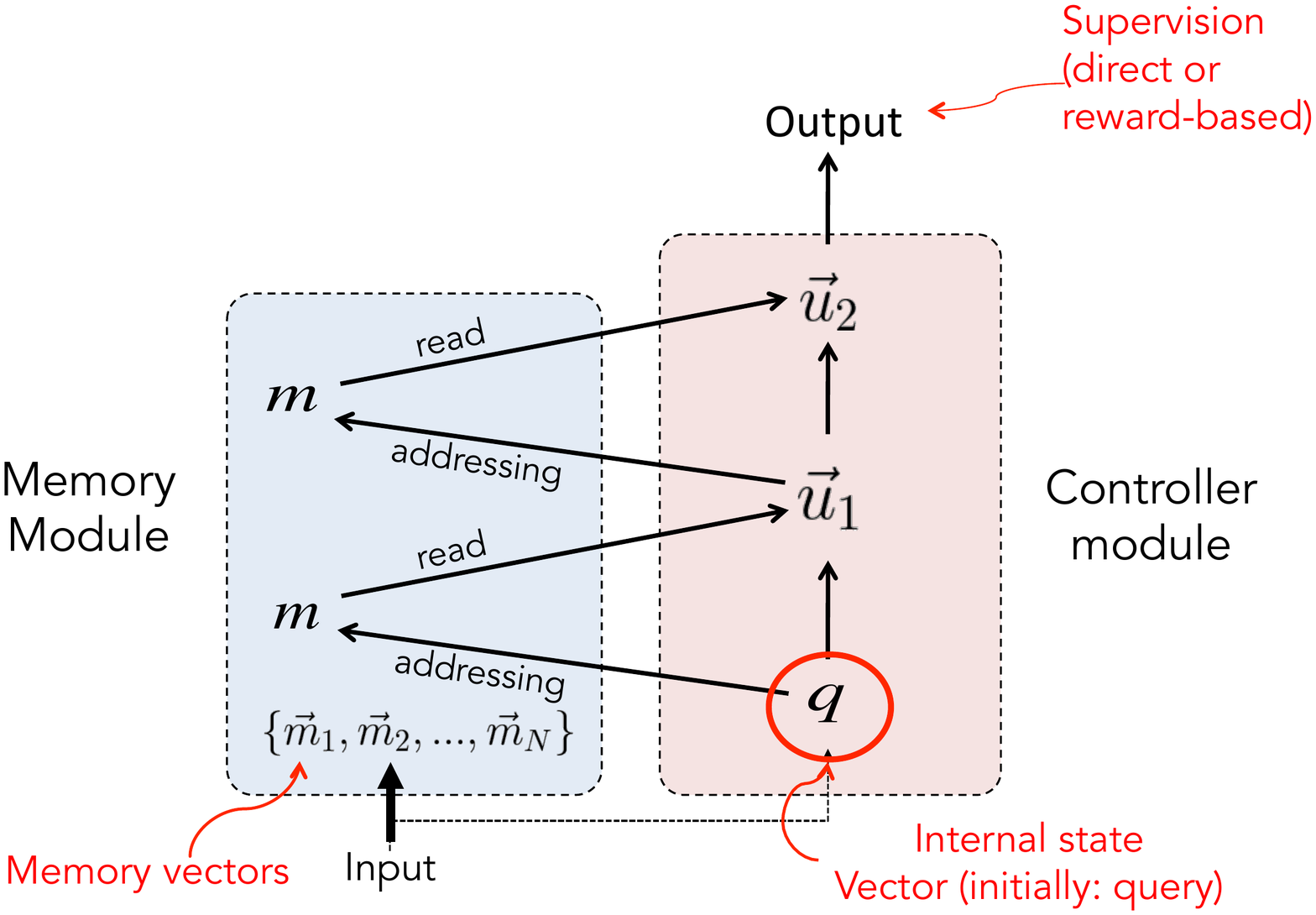}\includegraphics[width=7.3cm]{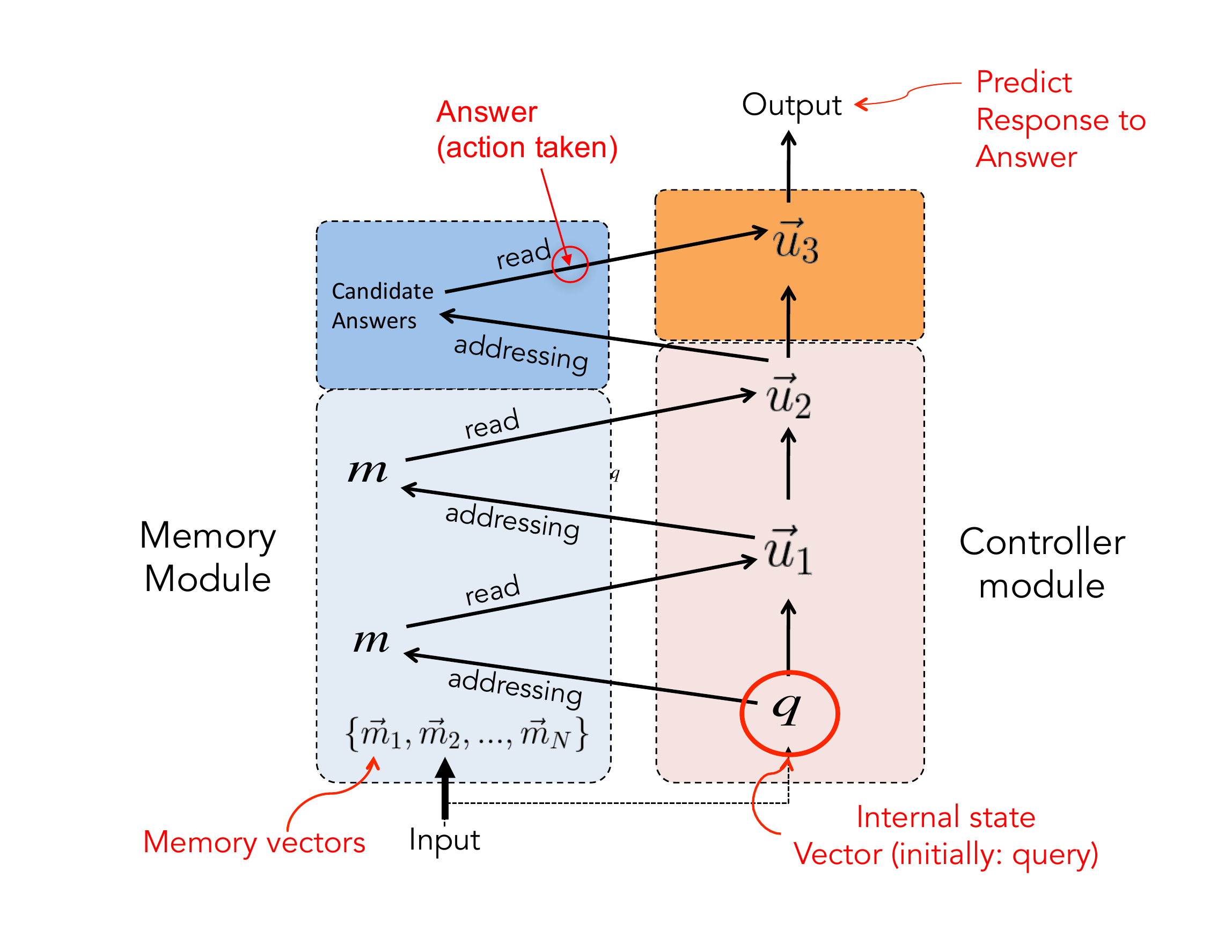}
(a) Model for (reward-based) imitation learning~~~~~~~~~~
(b) Model for forward prediction
\end{center}
\end{figure*}

\vspace*{-1ex}
\section{Learning Models}\label{sec:models}
\vspace*{-1ex}


Our main goal is to explore training strategies that can execute dialog-based
language learning. To this end we evaluate four possible strategies:
imitation learning, reward-based imitation, forward prediction, and a combination
of reward-based imitation and forward prediction. We will subsequently describe each in turn.

We test all of these approaches with the same model architecture: an end-to-end memory network (MemN2N)
\cite{sukhbaatar2015end}. 
Memory networks \cite{memnns,sukhbaatar2015end} 
are a recently introduced model that have been shown to do
well on a number of text understanding tasks, including question answering
 and dialog  \cite{dodge2015evaluating,bordes2015large}, 
language modeling \cite{sukhbaatar2015end} and sentence completion \cite{hill2015goldilocks}.
In particular, they outperform LSTMs and other baselines on the bAbI datasets
\cite{weston2015towards} which we employ with dialog-based 
learning modifications in Sec. \ref{sec:tasks}. 
They are hence a natural baseline model for us to use in order to
explore
differing modes of learning in our setup. 
In the following we will first review memory networks, detailing the explicit choices 
of architecture we made, and then show how they can be modified and applied
to 
 our setting of dialog-based language learning.

\paragraph{Memory Networks}

A high-level description of the memory network
 architecture we use is given in Fig. \ref{fig:archs} (a).
The input is the last utterance of the dialog, $x$, as well as a set of memories
(context)
 $(c_1, \dots, c_N)$ which can encode both short-term memory,
e.g. recent previous utterances and replies,
 and long-term memories, e.g. facts that could be useful for answering questions.
The context inputs $c_i$ are converted into vectors $m_i$ via embeddings and are stored
in the memory.
The goal is to produce an output $\hat{a}$ by 
processing the input $x$ and using that to
address and read from the memory, $m$, possibly multiple times,
 in order to form a coherent reply.
In the figure the memory is read twice, which is termed  multiple ``hops'' of attention.

In the first step, the input $x$ 
is embedded using a matrix $A$ of size $d \times V$ where $d$ is the embedding
dimension and $V$ is the size of the vocabulary, giving $q = A x$, where the input
$x$ is as a bag-of-words vector.
Each memory $c_i$ is embedded using the same matrix, giving $m_i = A c_i$.
The output of addressing and then reading from memory in the first hop is:
\[
  o_1 = \sum_i p^1_i m_i, ~~~p^1_i = \text{Softmax}(q^\top m_i).
\]
Here, the match between the input and the memories is computed
by taking the inner product followed by a softmax, yielding $p^1$,
giving a probability vector over the memories. The goal is to select memories relevant to
the last utterance $x$, i.e. the most relevant have large values of $p^1_i$.
The output memory representation $o_1$ is then constructed using the weighted sum of memories,
i.e. weighted by $p^1$.
The memory output is then added to the original input, $u_1 = R_1 (o_1 + q)$, to form the new state of the controller, where $R^1$ is a $d \times d$
rotation matrix\footnote{Optionally, different dictionaries can be used for inputs, memories
and outputs instead of being shared.}.
The attention over the memory can then be repeated using $u_1$ instead of $q$ as the addressing vector, yielding:
\[
      o_2 = \sum_i p^2_i m_i, ~~ p^2_i = \text{Softmax}(u_1^\top m_i),
\]
The controller state is updated again with  $u_2 = R_2 (o_2 + u_1)$,
where $R_2$ is another $d \times d$ matrix to be learnt.
In a two-hop model the final output is then defined as:
\begin{equation}\label{eq:a}
 \hat{a} = 
           \text{Softmax}(u_2^\top A y_1, \dots, u_2^\top A y_C)
\end{equation}
where there are $C$ candidate answers in $y$.
In our experiments $C$ is the set of actions that occur in the training set for the bAbI tasks,
and for MovieQA it is the set of words retrieved from the KB.

Having described the basic architecture, we now detail
the possible training strategies we can employ for our tasks.

\paragraph{Imitation Learning}
This approach involves simply imitating one of the speakers in observed dialogs,
which is essentially a supervised learning objective\footnote{Imitation learning algorithms are not always strictly supervised algorithms, they can also depend on the agent's actions. That is not the setting we use here, where the task is to imitate one of the speakers in a dialog.}.
This is the setting that most existing dialog learning, as well as question answer systems, 
employ for learning.
Examples arrive as $(x,c,a)$ triples, where $a$ is (assumed to be) a good response to 
the last utterance $x$ given context $c$.
In our case, the whole memory network model defined above
is trained using stochastic gradient 
descent by minimizing a standard cross-entropy
loss between $\hat{a}$ and the label $a$.

\paragraph{Reward-based Imitation}

If some actions are poor choices, then one does not want to repeat them,
that is we shouldn't treat them as a supervised objective.
In our setting positive reward is only obtained
immediately after (some of) the correct actions, or else is zero.
A simple strategy is thus to only apply imitation learning on the rewarded actions.
The rest of the actions are simply discarded from the training set.
This strategy is derived naturally as the degenerate
case one obtains by applying policy gradient \citep{williams1992simple} in our setting where the policy is fixed (see end of Sec. \ref{sec:tasks}).
In more complex settings (i.e. where actions that are made lead to long-term changes in the environment
and delayed rewards) applying reinforcement learning algorithms
  would be necessary, e.g. one could still use policy gradient to train the MemN2N but applied to the model's own policy, as used in \cite{DBLP:journals/corr/SukhbaatarSSCF15}.

\paragraph{Forward Prediction}

An alternative method of training is to perform forward prediction:
the aim is, given an utterance $x$ from speaker 1 and an answer $a$ by speaker 2 (i.e., the learner),
to predict $\bar{x}$,
the {\em{response to the answer}} from speaker 1. That is, in general to predict
the changed state of the world after action $a$, which in this case involves the
new utterance $\bar{x}$.

To learn from such data we propose the following modification to memory networks,
also shown in Fig. \ref{fig:archs} (b): essentially we chop off the final output from
the original network of Fig. \ref{fig:archs} (a) and replace it with some additional
layers that compute the forward prediction. The first part of the network remains
exactly the same and {\em only has access} to input $x$ and context $c$, just as before.
The computation up to $u_2 = R_2(o_2 + u_1)$ is thus exactly the same as before.

At this point we observe that the computation of the output in the original network,
by scoring candidate answers in eq. (\ref{eq:a}) looks similar to the addressing of
memory. Our key idea is thus to perform another ``hop'' of attention but over the candidate
answers rather than the memories. Crucially, we also
incorporate the information of which action (candidate)
was actually selected in the dialog (i.e. which one is $a$).
After this ``hop'', the resulting state of the controller is then used
to do the forward prediction.

Concretely, we compute:
\begin{equation} \label{eq:o3}
    o_3 =  \sum_i p^3_i (Ay_i + \beta^* [a = y_i]), ~~~~~ p^3_i = \text{Softmax}(u_2^\top A y_i),
\end{equation}
where $\beta^*$ is a $d$-dimensional vector, that is also learnt, that represents in the output $o_3$
the action that was actually selected.
After obtaining $o_3$, the forward prediction is then computed as:
\[
    \hat{x} = \text{Softmax}(u_3^\top A \bar{x}_1, \dots, u_3^\top A \bar{x}_{\bar{C}})
\]
where $u_3 = R_3 (o_3 + u_2)$.
That is, it computes the scores of the possible responses to the answer $a$ over
$\bar{C}$ possible candidates.
The mechanism in eq. (\ref{eq:o3}) gives the model a way to compare the most likely 
answers to $x$ with the given answer $a$, which in terms of supervision we believe is
critical. For example in question answering if the given answer $a$ is incorrect and the model
can assign high $p_i$ to the correct answer then the output $o_3$ will contain a small amount of 
 $\beta^*$; conversely, $o_3$ has a large amount of $\beta^*$ if $a$ is correct. Thus, $o_3$ informs the model of
the likely response $\bar{x}$ from the teacher.


Training can then be performed using the cross-entropy
loss between $\hat{x}$ and the label $\bar{x}$, similar to before.
In the event of a large number of candidates $\bar{C}$ we subsample the negatives,
always keeping  $\bar{x}$ in the set. The set of answers $y$ can also be similarly
sampled, 
making the method highly scalable.

A major benefit of this particular architectural design for forward prediction
 is that after training with the forward prediction criterion,
at test time one can ``chop off'' the top again of the model to retrieve the original
memory network  model of Fig. \ref{fig:archs} (a). One can thus use it to predict answers
$\hat{a}$ given only $x$ and $c$. 
We can thus evaluate its performance directly for that goal as well.

Finally, and importantly, 
if the answer to the response $\bar{x}$ carries pertinent supervision information
for choosing $\hat{a}$, as for example in many of the settings of Sec. \ref{sec:tasks} (and
Fig. \ref{taskseta}), then this will be backpropagated through the model.
This is simply not the case in the imitation, reward-shaping \cite{su2015reward}  or
reward-based imitation learning strategies which concentrate on the $x,a$ pairs.

\paragraph{Reward-based Imitation + Forward Prediction}

As our reward-based imitation learning uses the architecture of Fig. \ref{fig:archs} (a),
and forward prediction  uses the same architecture but with the additional layers
of Fig \ref{fig:archs} (b), we can learn jointly with both strategies.
One simply shares the weights across the two networks,
and performs gradient steps for both criteria, one of each type per action.
The former makes use of the reward signal -- which when available is a very useful signal --
but fails to use potential supervision feedback in the subsequent utterances, 
as described  above. It also effectively ignores dialogs carrying no reward.
Forward prediction in contrast makes use of dialog-based feedback and can train without any reward.
On the other hand not using rewards when available is a serious handicap.
Hence, the mixture of both strategies is a potentially powerful combination.

\begin{table*}[t!]
\caption{Test accuracy (\%) on the Single Supporting Fact bAbI dataset for various supervision approachess (training with 1000 examples on each) and different policies $\pi_{acc}$.
A task is successfully passed if $\geq 95\%$ accuracy is obtained (shown in blue).
\label{table:results1}}
\resizebox{1\linewidth}{!}{
 \begin{tabular}{lccc||ccc||ccc||ccc}
%
 & 
\multicolumn{3}{c|}{$\memn2n$} &
\multicolumn{3}{c|}{$\memn2n$} &
\multicolumn{3}{c|}{$\memn2n$} &
\multicolumn{3}{c}{ } \\
 &
\multicolumn{3}{c|}{\em imitation} &
\multicolumn{3}{c|}{\em reward-based} &
\multicolumn{3}{c|}{\em forward} &
\multicolumn{3}{c}{$\memn2n$} \\
 & 
\multicolumn{3}{c|}{\em learning} &
\multicolumn{3}{c|}{\em imitation (RBI)} &
\multicolumn{3}{c|}{\em prediction (FP)} &
\multicolumn{3}{c}{\em  RBI + FP} \\
\hline
     {\rotatebox[origin=l]{0}{~~~~~~~~~~~~~~Supervision Type~~~~~~~~~~~} $\pi_{acc}$~=}& 
     0.5 & 0.1 & 0.01 &
     0.5 & 0.1 & 0.01 &
     0.5 & 0.1 & 0.01 &
     0.5 & 0.1 & 0.01 \\
\hline
1 - Imitating an Expert Student    &  \PT{100}& \PT{100}  & \PT{100} 
                                   & \PT{100} & \PT{100} & \PT{100} 
                                   &  \FT{23} & \FT{30}  & \FT{29}    
                                   & \PT{99} & \PT{99}   & \PT{100}  \\
2 - Positive and Negative Feedback &  \FT{79} & \FT{28}  & \FT{21} 
                                   & \PT{99} & \FT{92}   & \FT{91} 
                                   &  \FT{93} & \FT{54}  & \FT{30} 
                                   & \PT{99} & \FT{92}   & \PT{96} \\
3 - Answers Supplied by Teacher    &  \FT{83} & \FT{37}  & \FT{25} 
                                   & \PT{99} & \PT{96}   & \FT{92} 
                                   &  \PT{99} & \PT{96}  & \PT{99} 
                                   & \PT{99} & \PT{100}  & \PT{98} \\
4 - Hints Supplied by Teacher      &   \FT{85} & \FT{23} & \FT{22} 
                                   & \PT{99} & \FT{91}   & \FT{90} 
                                   &  \PT{97} & \PT{99}  & \FT{66}     
                                   & \PT{99} & \PT{100}  & \PT{100} \\
5 - Supporting Facts Supplied by Teacher  & \FT{84} & \FT{24} & \FT{27} 
                                          & \PT{100} & \PT{96} & \FT{83}  
                                          & \PT{98} & \PT{99}& \PT{100}   
                                          & \PT{100} & \PT{99}& \PT{100}  \\
6 - Partial Feedback               & \FT{90} & \FT{22} & \FT{22}  
                                   & \PT{98} & \FT{81} & \FT{59} 
                                   & \PT{100} & \PT{100} & \PT{99}     
                                   & \PT{99} & \PT{100} & \PT{99} \\
7 - No Feedback                    & \FT{90} & \FT{34} & \FT{19}  
                                   & \FT{20} & \FT{22} & \FT{29} 
                                   & \PT{100} & \PT{98} & \PT{99}    
                                   & \PT{98} & \PT{99} & \PT{99}  \\
 8 - Imitation + Feedback Mixture  & \FT{90} & \FT{89} & \FT{82}  
                                   & \PT{99} & \PT{98} & \PT{98} 
                                   & \FT{28} & \FT{64} & \FT{67}    
                                   & \PT{99} & \PT{98} & \PT{97} \\
 9 - Asking For Corrections        & \FT{85} & \FT{30} & \FT{22}  
                                   & \PT{99} & \FT{89} & \FT{83} 
                                   & \FT{23} & \FT{15} & \FT{21}    
                                   & \PT{95} & \FT{90} & \FT{84} \\
 10 - Asking For Supporting Facts  & \FT{86} & \FT{25} & \FT{26}  
                                   & \PT{99} & \PT{96} & \FT{84} 
                                   & \FT{23} & \FT{30} & \FT{48}    
                                   & \PT{97} & \PT{95} & \FT{91}  \\
\cline{1-13}
 Number of completed tasks $(\geq95\%)$    & 1 & 1 & 1 & 9 & 5 & 2 & 5 & 5 & 4 & 10 & 8 & 8 \\ 
\end{tabular}
} 
\vspace*{-3ex}
\end{table*}

\begin{table*}[h!]
\caption{Test accuracy (\%) on the MovieQA dataset
 dataset for various supervision approaches. Numbers in bold are the winners for that task and
choice of $\pi_{acc}$.
\label{table:results2}}
\resizebox{1\linewidth}{!}{
 \begin{tabular}{lccc||ccc||ccc||ccc}
%
%
 & 
\multicolumn{3}{c|}{$\memn2n$} &
\multicolumn{3}{c|}{$\memn2n$} &
\multicolumn{3}{c|}{$\memn2n$} &
\multicolumn{3}{c}{} \\
 &
\multicolumn{3}{c|}{\em imitation} &
\multicolumn{3}{c|}{\em reward-based} &
\multicolumn{3}{c|}{\em forward} &
\multicolumn{3}{c}{$\memn2n$} \\
 & 
\multicolumn{3}{c|}{\em learning} &
\multicolumn{3}{c|}{\em imitation (RBI)} &
\multicolumn{3}{c|}{\em prediction (FP)} &
\multicolumn{3}{c}{\em  RBI + FP} \\
\hline
     {\rotatebox[origin=l]{0}{~~~~~~~~~~~~~~Supervision Type~~~~~~~~~~~} $\pi_{acc}$~=}& 
     0.5 & 0.1 & 0.01 &
     0.5 & 0.1 & 0.01 &
     0.5 & 0.1 & 0.01 &
     0.5 & 0.1 & 0.01 \\
\hline
1 - Imitating an Expert Student    & \FT{\bf 80}& \FT{\bf80}  & \FT{\bf 80} 
                                   & \FT{80}& \FT{80}  & \FT{80} 
                                   & \FT{24}& \FT{23}  & \FT{24} 
                                   & \FT{77}& \FT{77}  & \FT{77}  \\
2 - Positive and Negative Feedback & \FT{46}& \FT{29}  & \FT{27} 
                                   & \FT{52}& \FT{32}  & \FT{26} 
                                   & \FT{48}& \FT{34}  & \FT{24} 
                                   & \FT{\bf 68}& \FT{\bf 53} & \FT{\bf 34}  \\
3 - Answers Supplied by Teacher    & \FT{48}& \FT{29}  & \FT{26} 
                                   & \FT{52}& \FT{32}  & \FT{27} 
                                   & \FT{60}& \FT{57} & \FT{58} 
                                   & \FT{\bf 69}& \FT{\bf 65}  & \FT{\bf 62}  \\
4 - Hints Supplied by Teacher      & \FT{47}& \FT{29}  & \FT{26} 
                                   & \FT{51}& \FT{32}  & \FT{28} 
                                   & \FT{58}& \FT{\bf 58}  & \FT{\bf 42} 
                                   & \FT{\bf 70}& \FT{54}  & \FT{32}  \\
5 - Supporting Facts Supplied by Teacher &
                                      \FT{47}& \FT{28}  & \FT{26} 
                                   &  \FT{51}& \FT{32}  & \FT{26}
                                   & \FT{43}& \FT{44}  & \FT{33} 
                                   & \FT{\bf 66}& \FT{\bf 53}  & \FT{\bf 40}  \\
6 - Partial Feedback               & \FT{48}& \FT{29}  & \FT{27} 
                                   & \FT{49}& \FT{32}  & \FT{24} 
                                   & \FT{60}& \FT{58}  & \FT{58} 
                                   & \FT{\bf 70}& \FT{\bf 63}  & \FT{\bf 62}  \\
7 - No Feedback                    & \FT{51}& \FT{29}  & \FT{27} 
                                   & \FT{22}& \FT{21}  & \FT{21} 
                                   & \FT{60}& \FT{53}  & \FT{\bf 58} 
                                   & \FT{\bf 61}& \FT{\bf 56}  & \FT{50}  \\
 8 - Imitation + Feedback Mixture & \FT{60}& \FT{50}  & \FT{47} 
                                   & \FT{63}& \FT{53}  & \FT{51} 
                                   & \FT{46}& \FT{31}  & \FT{23} 
                                   & \FT{\bf 72}& \FT{\bf 69}  & \FT{\bf 69}  \\
 9 - Asking For Corrections        & \FT{48}& \FT{29} & \FT{27} 
                                   & \FT{52}& \FT{34}  & \FT{26} 
                                   & \FT{67}& \FT{\bf 52}  & \FT{\bf 44} 
                                   & \FT{\bf 68}& \FT{\bf 52}  & \FT{39}  \\
 10 - Asking For Supporting Facts  & \FT{49}& \FT{29}  & \FT{27} 
                                   & \FT{52}& \FT{34}  & \FT{27} 
                                   & \FT{51}& \FT{44}  & \FT{35} 
                                   & \FT{\bf 69} & \FT{\bf 53}  & \FT{\bf 36}  \\
\cline{1-13}        
 Mean Accuracy                      & \FT{52}& \FT{36}  & \FT{34} 
                                   & \FT{52}& \FT{38}  & \FT{34} 
                                   & \FT{52}& \FT{45}  & \FT{40} 
                                   & \FT{\bf 69}& \FT{\bf 60}  & \FT{\bf 50}  \\
\end{tabular}
}
\vspace*{-3ex}
\end{table*}

\section{Experiments} \label{sec:exp}

We conducted experiments on the datasets described in Section \ref{sec:tasks}.
As described before, for each task we consider a fixed policy for performing actions
(answering questions) which gets questions correct with probability $\pi_{acc}$. 
We can thus compare the different training strategies described in Sec. \ref{sec:models}
over each task for different values of $\pi_{acc}$. Hyperparameters for all methods are optimized
on the validation sets.
A summary of the results is reported in Table \ref{table:results1} for the bAbI dataset and 
Table \ref{table:results2} for MovieQA.
We observed the following results:
\begin{itemize}
\item Imitation learning, ignoring rewards, is a poor learning strategy when
imitating inaccurate answers, e.g. for $\pi_{acc}< 0.5$. For imitating an expert however (Task 1) it is hard to beat.
\item Reward-based imitation (RBI) performs better when rewards are available,
particularly in Table \ref{table:results1}, but also degrades when they are
too sparse e.g. for  $\pi_{acc}=0.01$.
\item Forward prediction (FP) is  more robust and has stable performance at different levels of  $\pi_{acc}$. However
as it only predicts answers implicitly and does not make use of rewards it is outperformed by RBI on several tasks, notably Tasks 1 and 8 (because it cannot do supervised learning) and Task 2 
(because it does not take advantage of positive rewards).
\item  FP makes use of dialog feedback in Tasks 3-5 whereas RBI does not. 
This explains why FP does better with useful feedback (Tasks 3-5) than without (Task 2), whereas RBI cannot.
\item Supplying full answers (Task 3) is more useful than hints (Task 4) but hints still help FP more than just yes/no answers without extra information (Task 2). 
\item When positive feedback is sometimes missing (Task 6) RBI suffers especially in Table \ref{table:results1}. FP does not as it does not use this feedback.
\item One of the most surprising results of our experiments is that FP performs well overall, given that it does not use feedback, which we will attempt to explain subsequently.
This is particularly evident on Task 7 (no feedback) where
RBI has no hope of succeeding as it has no positive examples. FP on the other hand learns adequately.
\item Tasks 9 and 10 are harder for FP as the question is not immediately before the feedback.
\item Combining RBI and FP ameliorates the failings of each, yielding the best overall results.
\end{itemize}

One of the most interesting aspects of our results is that FP works at all without any rewards.
In Task 2 it does not even ``know'' the difference between words like ``yes'' or ``'correct'' vs. 
words like ``wrong'' or ``incorrect'', so why should it tend to predict actions that lead to a response like ``yes, that's right''? 
This is because there is a natural coherence to predicting true answers that leads to greater accuracy in forward prediction. That is, you cannot predict a ``right'' or ``wrong'' response from the teacher 
if you don't know what the right answer is.
 In our experiments our policies $\pi_{acc}$ sample negative answers equally, which may make learning simpler.
We thus conducted an experiment on Task 2 (positive and negative feedback) of the bAbI dataset with a much more biased policy: it is the same as $\pi_{acc}=0.5$ except when the policy predicts incorrectly there is probability 0.5 of choosing a random guess as before, and 0.5 of choosing the fixed answer {\em bathroom}.
In this case the FP method obtains 68\% accuracy showing the method still works in this regime, 
although not as well as before.

\section{Conclusion}

We have presented a set of evaluation datasets and models for dialog-based language learning.
The ultimate goal of this line of research is to move towards a learner 
capable of talking to humans, such that
humans are able to effectively teach it during dialog.
We believe the dialog-based language learning approach we described is
a small step towards that goal.

This paper only studies some restricted types of feedback,
namely positive feedback and corrections of various types.
However, potentially any reply in a dialog can be seen as feedback, and should be useful for learning. It should be studied if forward prediction, and the other approaches we tried,  work there too.
Future work should also develop further evaluation methodologies to test 
how the models we presented here, and new ones, work in those settings, e.g. 
in more complex settings where actions that are made lead to long-term changes in
the environment and delayed rewards, i.e. extending to the reinforcement learning setting.
Finally, dialog-based feedback could also be used as a medium to learn non-dialog based skills,
e.g. natural language dialog for completing visual or physical tasks.

\section*{Acknowledgments} 
We thank  Arthur Szlam, Y-Lan Boureau, Marc'Aurelio Ranzato, Ronan Collobert, 
Michael Auli, David Grangier,
Alexander Miller, Sumit Chopra, Antoine Bordes and Leon Bottou
for helpful discussions and feedback, and the Facebook AI Research
team in general for supporting this work.

\small
\bibliographystyle{abbrv}
\bibliography{refs}

\end{document}